\documentclass{article}

\usepackage[nonatbib,preprint]{neurips_2024}
\usepackage{enumitem}
\usepackage[utf8]{inputenc} 
\usepackage[T1]{fontenc}    
\usepackage{hyperref}       
\usepackage{url}            
\usepackage{booktabs}       
\usepackage{amsfonts}       
\usepackage{nicefrac}       
\usepackage{microtype}      
\usepackage{xcolor}         
\usepackage{graphicx}
\usepackage{multirow}
\usepackage{amssymb}
\usepackage[normalem]{ulem}
\useunder{\uline}{\ul}{}

\title{ID-Animator: Zero-Shot Identity-Preserving Human Video Generation}

\author{Xuanhua He$^{1,2}$\footnotemark[1], Quande Liu$^{3}$\footnotemark[2], Shengju Qian$^{3}$, Xin Wang$^{3}$, \\ \textbf{Tao Hu}$^{1,2}$, \textbf{Ke Cao}$^{1,2}$, \textbf{Keyu Yan}$^{1,2}$, \textbf{Jie Zhang}$^{2}$\footnotemark[2]\\ 
{\tt\small$^{1}$University of Science and Technology of China}\\
{\tt\small$^{2}$Hefei Institute of Physical Science, Chinese Academy of Sciences}\\
{\tt\small$^{3}$LightSpeed Studios, Tencent}\\
 \vspace{5mm}
{\tt\small \url{https://id-animator.github.io/}}
}

\begin{document}

\maketitle
\let\thefootnote\relax\footnotetext{$^*$ Intern in Tencent \hspace{3pt} $^\dagger$ Corresponding authors
}
\begin{figure}[!ht]
    \centering
    \includegraphics[width=\linewidth]{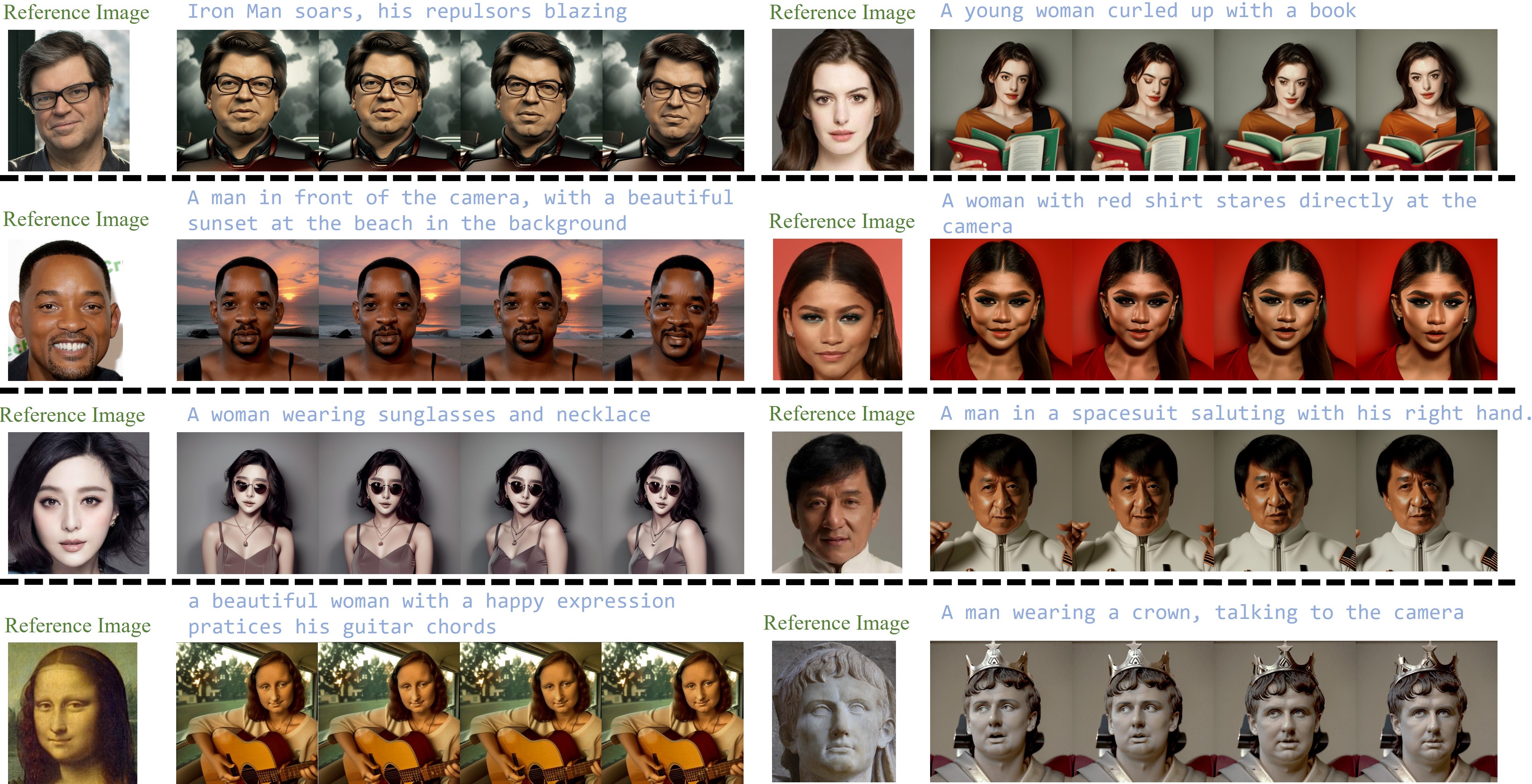}
    \caption{Given simply one facial image, our \textbf{ID-Animator} is able to produce a wide range of personalized videos that not only preserve the identity of input image, but further align with the given text prompt, all within a single forward pass \textbf{without further tuning}.}
    \label{fig:enter-label}
\end{figure}
\begin{abstract}
Generating high-fidelity human video with specified identities has attracted significant attentions in the content generation community. However, existing techniques struggle to strike a balance between training efficiency and identity preservation, either requiring tedious case-by-case fine-tuning or usually missing identity details in the video generation process. 
In this study, we present \textbf{ID-Animator}, a zero-shot human-video generation approach that can perform personalized video generation given a single reference facial image without further training. ID-Animator inherits existing diffusion-based video generation backbones with a face adapter to encode the ID-relevant embeddings from learnable facial latent queries. To facilitate the extraction of identity information in video generation, we introduce an ID-oriented dataset construction pipeline that incorporates unified human attributes and action captioning techniques from a constructed facial image pool. Based on this pipeline, a random reference training strategy is further devised to precisely capture the ID-relevant embeddings with an ID-preserving loss, thus improving the fidelity and generalization capacity of our model for ID-specific video generation. Extensive experiments demonstrate the superiority of ID-Animator to generate personalized human videos over previous models. Moreover, our method is highly compatible with popular pre-trained T2V models like animatediff and various community backbone models, showing high extendability in real-world applications for video generation where identity preservation is highly desired. Our codes and checkpoints are released.
\end{abstract}

\section{Introduction}
Personalized or customized generation is to create visual content consistent in style, subject, or character ID based on one or more reference images.
In the realm of image generation, considerable strides have been made in crafting this identity-specific content, particularly in the domain of human image synthesis~\cite{li2023photomaker,valevski2023face0,yan2023facestudio,wang2024instantid}. Recently, text-driven video generation~\cite{guo2023animatediff,wang2023modelscope,wu2023tune} has gathered substantial interest within the research community. These methods enable the creation of videos based on user-specified textual prompts. However, the quest for generating high-fidelity, identity-specific human videos remains to be explored. The generation of identity-specific human videos holds profound significance, particularly within the film, advertisement and game industries, etc, where usually a representative character is required to appear. Previous approaches for customization primarily emphasized specified postures~\cite{hu2023animate}, styles~\cite{liu2023stylecrafter}, and action sequences~\cite{xu2024you}, often employing additional control to ensure that the generated videos met user requirements. However, these methods largely overlook the controllability  on  identity information. Some techniques involved model fine-tuning through methods like LoRA~\cite{hu2021lora} and textural inversion~\cite{gal2022image} to achieve ID-specific control~\cite{ma2024magic}, but requiring tedious case-by-case training for each ID thus lacking real-time inference ability. Others relied on image prompts to guide the model to feature particular subjects in generating videos, yet encountering challenges such as intricate dataset pipeline construction and limited ID variations~\cite{jiang2023videobooth}. Furthermore, the direct integration of image customization modules~\cite{ye2023ip} into the video generation model resulted in poor quality, such as static motion or severe frame inconsistency. 

In summary, the field of ID-specified video generation currently confronts several notable challenges:
\begin{enumerate}
    \item \textbf{High training and fine-tuning costs}: Many ID-specified customization methods requires tedious case-by-case fine-tuning costs due to the the lack of enough prior knowledge, consequently imposing significant training overheads at inference time. These training costs hinder the widespread adoption and scalability of ID-specified video generation techniques.
    \item \textbf{Scarcity of high-quality text-conditioned human video datasets}: Unlike the image generation community, where datasets like LAION-face~\cite{zheng2022general} are readily available, the video generation community lacks sufficient high-quality text-video data pairs, especially for human videos. Existing datasets, such as CelebV-text~\cite{yu2023celebv}, feature captions annotated with fixed templates that concentrate on emotion changes while ignoring human attributes and actions, making them unsuitable for ID-preserving video generation tasks. This scarcity hampers research progress, leading to repetitive endeavors to collect private datasets.
    \item \textbf{Influence of ID-irrelevant features from reference images for video generation}: The existence of ID-irrelevant features in reference images can adversely hurt the quality and identity preservation of generated videos. How to reduce the influence of such features poses a challenge, demanding novel solutions to ensure fidelity in ID-specified video generation.
\end{enumerate}
\textbf{Solutions:}
To address the above issues, we propose an efficient ID-specific video generation framework, named ID-Animator. With the help of the pre-trained text-to-video diffusion model and a lightweight face adapter to encode the ID-relevant embeddings from learnable facial latent queries, our method can largely reduce the training and fine-tuning costs for the first issue, i.e., it can complete training within one day on a single A100 GPU and perform zero-shot inference once the training is done.
To address the second issue, we build an ID-oriented dataset construction pipeline. By leveraging existing publicly available datasets, we introduce the concept of unified captions, which involves generating captions for human actions, human attributes and a unified human description. Additionally, we leverage face-relevant recognition techniques to create corresponding reference image pool. 
Trained with the rewritten captions and the extracted facial image pool, 
our ID-Animator significantly enhances its video generation quality.
In response to the third issue, we devise a novel random reference training strategy, which randomly samples faces from the face pool and optimize with an ID-preserving objective to decouple ID-independent 
content from ID-related facial features.


Through the aforementioned designs, our model can achieve zero-shot ID-specific video generation in a lightweight manner. It seamlessly integrates into existing community models~\cite{civitai}, showcasing robust generalization and ID preservation capabilities.

Our core contribution can be summarized as follows:
\begin{itemize}
    \item We propose ID-Animator, a novel framework that can generate identity-specific videos given any reference facial image without further model fine-tuning. It inherits pre-trained video diffusion models with a lightweight face adapter to encode the ID-relevant embeddings from learnable facial latent queries. To the best of our knowledge, this is the first endeavor towards achieving zero-shot ID-specific human video generation.
    \item We develop an ID-oriented dataset construction pipeline to mitigate the missing of training dataset in personalized human video generation. Over publicly available data sources, we present unified captioning of human videos, which extracts textual descriptions for human attributes and actions respectively to attain comprehensive human captions. Besides, a facial image pool is constructed over this dataset to facilitate the extraction of facial embeddings. 
    \item Over this pipeline, we further devise a random reference training strategy optimized with an ID-preserving objective, aiming to precisely extract the identity-relevant features and diminish the influence of ID-irrelevant information from the given reference facial image, thereby improving the identity fidelity and generalization ability of ID-Animator in real-world applications for personalized human video generation.
\end{itemize}
\section{Related Work}
\subsection{Video Generation}
Video generation has been a key area of interest in research for a long time. 
Early endeavors in the task utilized models like generative adversarial networks~\cite{hong2022depth,chu2020learning,hu2022make} and vector quantized variational autoencoder to generate video~\cite{jiang2023text2performer,ge2022long,qian2023strait}. However, due to the inherent model ability, this video lacks motion and details and is unable to achieve good results.
With the rise of the diffusion model~\cite{ho2020denoising}, notably the latent diffusion model~\cite{rombach2022high} and its success in image generation, researchers have extended the diffusion model's applicability to video generation~\cite{ho2022video,ho2022imagen,blattmann2023stable}. 
This technique can be classified into two parts: image-to-video and text-to-video generation.
The former essentially transforms a given image into a dynamic video, whereas the latter generates video only following text instructions, without any image as input.
Leading-edge methods, exemplified by these works, include Animate Diffusion~\cite{guo2023animatediff}, Dynamicrafter~\cite{xing2023dynamicrafter}, Modelscope~\cite{wang2023modelscope}, AnimateAnything~\cite{dai2023animateanything}, and Stable Video~\cite{blattmann2023stable}, among others. These techniques generally exploit pre-trained text-to-image models and intersperse them with diverse forms of temporal mixing layers. 

\subsection{ID-preserving Image Generation}
The impressive generative abilities of diffusion models have attracted recent research endeavors investigating their personalized generation potential. Current methods within this domain can be divided into two categories, based on the necessity of fine-tuning during the testing phase. A subset of these methods requires the fine-tuning of the diffusion model leveraging ID-specific datasets during the testing phase, representative techniques such as DreamBooth~\cite{ruiz2023dreambooth}, textual inversion~\cite{gal2022image}, and LoRA~\cite{hu2021lora}. 
While these methods exhibit acceptable ID preservation abilities, they necessitate individual model training for each unique ID, thus posing a significant challenge related to training costs and dataset collection, subsequently hindering their practical applicability. 
The latest focus of research in this domain has shifted towards training-free methods that bypass additional fine-tuning or inversion processes in testing phase. During the inference phase, it is possible to create a high-quality ID-preserving image with just a reference image as the condition. Methods like Face0~\cite{valevski2023face0} replace the final three tokens of text embedding with face embedding within CLIP's feature space, utilizing this new embedding as conditional for image generation. PhotoMaker~\cite{li2023photomaker}, on the other hand, takes a similar approach by stacking multiple images to reduce the influence of ID-irrelevant features. Similarly, IP-Adapter~\cite{ye2023ip} decoupled reference image features and text features to facilitate cross attention, resulting in better instruction following. 
Concurrently, InstantID~\cite{wang2024instantid} combined the features of IP-Adapter and ControlNet~\cite{zhang2023adding}, utilizing both global structural attributes and the fine-grained features of reference images for the generation of ID-preserving images. 
\subsection{Subject-driven Video Generation}
Subject-driven video generation aims at generating videos containing customized subjects.
Research on subject-driven video generation is still in its early stages, with two notable works being VideoBooth~\cite{jiang2023videobooth} and MagicMe~\cite{ma2024magic}. VideoBooth~\cite{jiang2023videobooth} strives to generate videos that maintain high consistency with the input subject by utilizing the subject's clip feature and latent embedding obtained through a VAE encoder. This approach offers more fine-grained information than ID-preserving generation methods; however, its limitation remains the subjects required to be present in the training data, such as cats, dogs, and vehicles, which results in a restricted range of applicable subjects. MagicMe~\cite{ma2024magic}, on the other hand, is more closely related to the ID-preserving generation task. It learns ID-related representations by generating unique prompt tokens for each ID. However, this method requires separate training for each ID, making it unable to achieve zero-shot training-free capabilities. This limitation poses a challenge for its practical application. Our proposed method distinguishes itself from these two approaches by being applicable to any human image without necessitating retraining during inference. 
\section{Method}
\begin{figure}[!h]
    \centering
    \includegraphics[width=\linewidth]{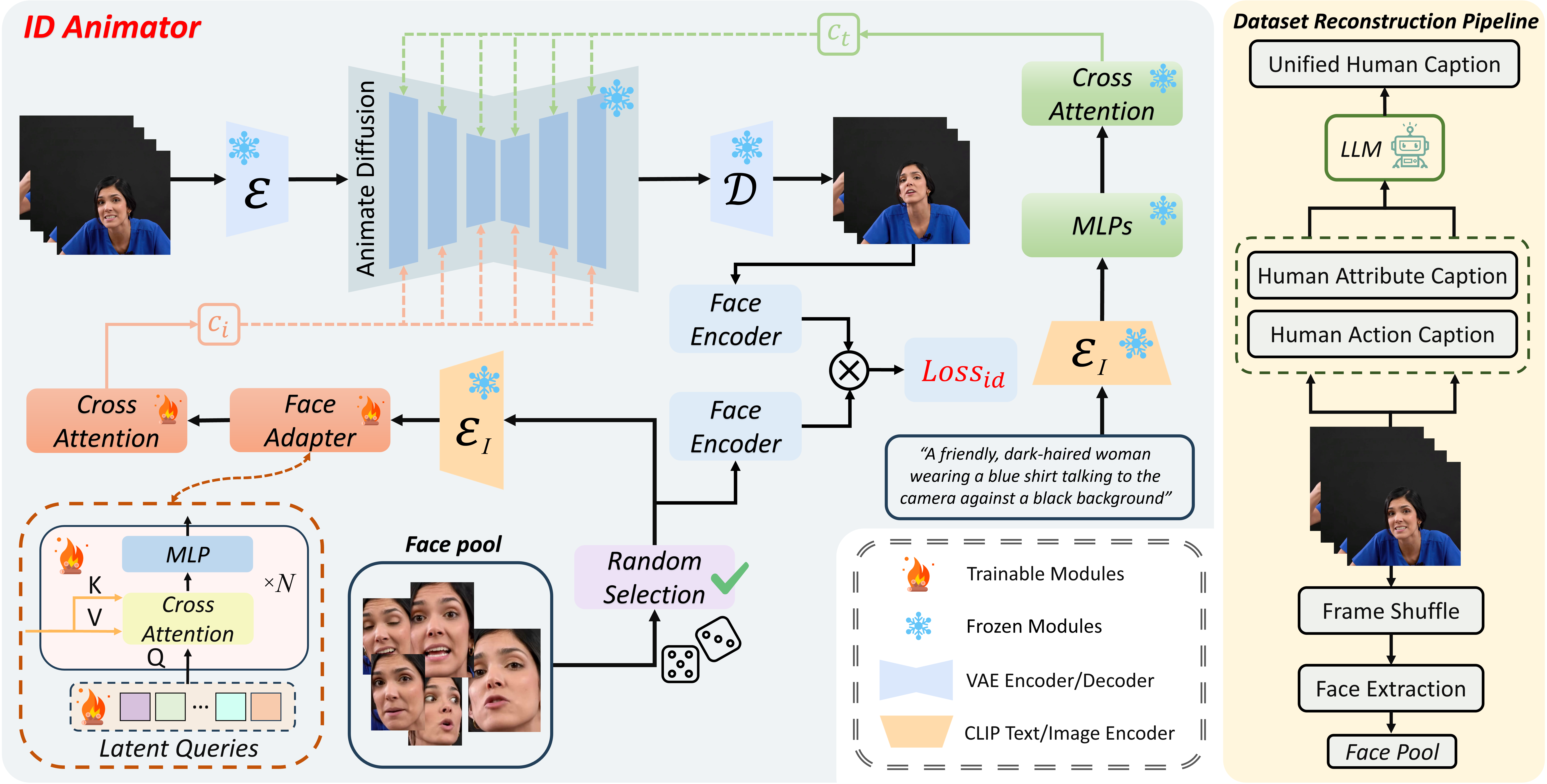}
    \caption{An overview of our proposed framework: the ID-Animator; dataset reconstruction pipeline; random reference training strategy with ID-preserving learning objective.}
    \label{fig:mainfig}
\end{figure}
\subsection{Overview}
Given a reference ID image, ID-Animator endeavors to produce high-fidelity ID-specific human videos. Figure~\ref{fig:mainfig} demonstrates our methods, featuring three pivotal constituents: a dataset reconstruction pipeline, the ID-Animator framework, and the random reference decompositional strategy employed during the training process of ID-Animator.
\subsection{ID-Animator}
As depicted at the bottom of Figure~\ref{fig:mainfig}, our ID-Animator framework comprises two components: the backbone text-to-video model, which is compatible with diverse T2V models, and the face adapter, which is subject to training for efficiency.

\noindent\textbf{Pretrained Text to Video Diffusion Model} The pre-trained text-to-video diffusion model exhibits strong video generation abilities, yet it lacks efficacy in the realm of ID-specific human video generation. Thus, our objective is to harness the existing capabilities of the T2V model and tailor it to the ID-specific human video generation domain. Specifically, we employ AnimateDiff~\cite{hu2023animate} as our foundational T2V model.

\noindent\textbf{Face Adapter}
The advent of image prompting has substantially bolstered the generative ability of diffusion models, particularly when the desired content is challenging to describe precisely in text. IP-Adapter~\cite{ye2023ip} proposed a novel method, enabling image prompting capabilities on par with text prompts, without necessitating any modification to the original diffusion model. Our approach mirrors the decoupling of image and text features in cross-attention. 
This procedure can be mathematically expressed as:
\begin{equation}
        Z_{new} = Attention(Q,K^{t},V^{t})+\lambda\cdot Attention(Q,K^{i},V^{i})
\end{equation}
where $Q$, $K^t$, and $V^t$ denote the query, key, and value matrices for text cross-attention, respectively, while $K^i$ and $V^i$ correspond to image cross-attention. Provided the query features $Z$ and the image features $c_i$, $Q = ZW_q$, $K^{i} = c_iW^{i}_k$, and $V^i =c_iW^{i}_{v}$. Only $W^{i}_{k}$ and $W^{i}_{v}$ are trainable weights.

Inspired by IP-Adapter, we limit our modifications to the cross-attention layer in the video generation model, leaving the temporal attention layer unchanged to preserve the original generative capacity of the model. 
A lightweight face adapter module is designed, encompassing a handful of simple query-based image encoder and the cross-attention module with trainable cross-attention projection weights, as shown in Figure~\ref{fig:mainfig}.
The image feature $c_i$ is derived from the clip feature of the reference image, and is further refined by the query-based image encoder. The other weights in cross attention module are initialized from the original diffusion model, of which the projection weights $W_K^{i}$ and $W_V^{i}$ are initialized using the weights of the IP-Adapter, facilitating the acquisition of preliminary image prompting capabilities and reducing the overall training costs.

\subsection{ID-Oriented Human Dataset Reconstruction}
Contrary to identity-preserving image generation tasks, video generation tasks currently suffer from a lack of identity-oriented datasets. The dataset most relevant to our work is the CelebV-HQ~\cite{yu2023celebv} dataset, comprising 35,666 video clips that encompass 15,653 identities and 83 manually labeled facial attributes covering appearance, action, and emotion. However, their captions are derived from manually set templates, primarily focusing on facial appearance and human emotion while neglecting the comprehensive environment, human action, and detailed attributes of video. Additionally, its style significantly deviates from the user instructions, rendering it unsuitable for contemporary video generation models. Consequently, we find it necessary to reconstruct this dataset into an identity-oriented human dataset. Our pipeline incorporates a unified caption technique and the construction of a face image pool.
\begin{figure}
    \centering
\includegraphics[width=\linewidth]{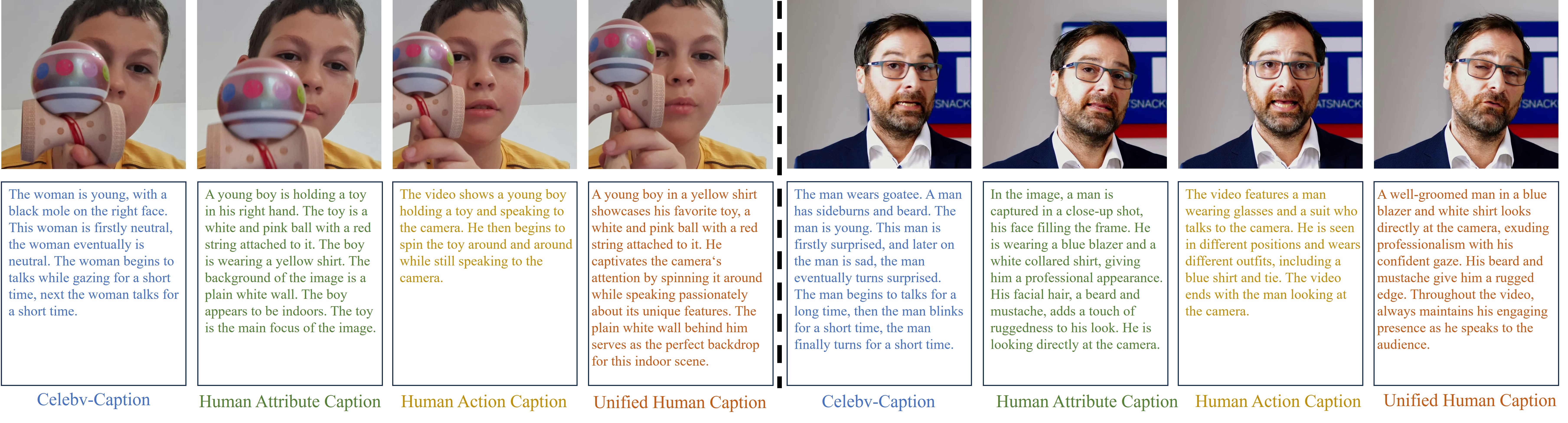}
    \caption{Examples of the original CelebV-caption and our human attribute caption, human action caption and the unified human caption.}
    \label{fig:caption}
\end{figure}

\noindent\textbf{Unified Human Video Caption Generation.}
We design a comprehensive restructuring of the captions within the CelebV-HQ dataset to enhance human video geneartion for ID-Animator. 
To produce high-quality human videos, it is crucial to comprehensively caption the semantic information and intricate details present within the video. Consequently, the caption must incorporate detailed attributes of the individual as well as the actions they are performing in the video. 
In light of this, we employ a novel rewriting technique that decouples the caption into two distinct components: human attributes and human actions. Subsequently, we leverage a language model to amalgamate these elements into a cohesive and comprehensive caption, as illustrated at the right of Figure~\ref{fig:mainfig}.

\begin{enumerate}[leftmargin = 15pt]
\item[$\bullet$] \textbf{Human Attribute Caption.}
As a preliminary step, we focus on crafting an attribute caption that aims to vividly depict the individual's appearance and the surrounding context. To achieve this, we employ the ShareGPT4V~\cite{chen2023sharegpt4v} model for caption generation.
We choose the median frame of the video as the input for ShareGPT4V. This approach allows us to generate detailed character descriptions that incorporate a wealth of attribute information.
\item[$\bullet$] \textbf{Human Action Caption.}
Our objective is to create human videos with accurate and rich motions, where a mere human attribute caption is insufficient for our needs. To address this requirement, we introduce the concept of a human action caption, which strives to depict the action present within the video. These captions are specifically designed to concentrate on the semantic content across the entire video, facilitating a comprehensive understanding of the individual's actions captured therein. To achieve this goal, we leverage the Video-LLava~\cite{lin2023video} model, which has been trained on video data and excels at focusing on the overall dynamism.
\item[$\bullet$] \textbf{Unified Human Caption.}
The limitations of relying solely on human attribute captions and human action captions are demonstrated in Figure~\ref{fig:caption}. Human attribute caption fails to encompass the overall action of the individual, while human action caption neglects the detailed characteristics of the subject. To address this, we designed a unified human caption that amalgamates the benefits of both caption types, using this comprehensive caption to train our model. We employ a large language model to facilitate this integration, capitalizing on its capacity for human-like expression and its capacities in generating high-quality captions. The GPT-3.5 API is utilized in this process. As depicted in Figure~\ref{fig:caption}, the rewritten caption effectively encapsulates the video scene, aligning more closely with human instructions.
\end{enumerate}
\noindent\textbf{Random Face Extraction for Face Pool Construction.}
In contrast to previous methods~\cite{ye2023ip,ma2024magic,jiang2023videobooth}, our approach does not directly utilize a frame from video as a reference image. Instead, we opt to extract the facial region from video, using this as the reference image.
Simultaneously, our technique differs from the image reconstruction training strategy
 employed in the ID-preserving image generation works~\cite{wang2024instantid,valevski2023face0,li2023photomaker}, which typically reconstructs a reference image using the same image as condition. As depicted at the bottom of Figure~\ref{fig:mainfig}, we employ shuffling on video sequences and extract facial regions from five randomly selected frames. In instances where a frame contains more than one face, it is discarded and additional frames are selected for re-extraction. The extracted facial images are subsequently stored in the face pool.

\subsection{Random Reference Training Methods For Diminishing ID-Irrelevant Features}

In the training phase of the ID-preserving diffusion model, the objective is to estimate the noise $\epsilon$ at the current time step $t$ from a noisy latent representation $z_t$, incorporating conditions such as a text condition $C$ and an image condition $C_i$. This noisy latent $z_t$ is derived from the clean latent $z$ combined with the noise component associated with the current time step $t$, i.e., $z_t=f (z, t)$. This optimization procedure can be expressed by the following function:
\begin{equation}
    \mathcal{L}_{diff} = \mathrm{E}_{z_t,t,C,C_i,\epsilon~\mathcal{N}(0,1)}[||\epsilon-\epsilon_{\theta}(z_{t},t,C,C_i)||_{2}^{2}]\label{eq}
\end{equation}
In current identity-preserving image generation models, the image condition $C_i$ and the reconstruction target $z$ typically originate from the same image $I$. For instance, Face0~\cite{valevski2023face0}, InstantID~\cite{wang2024instantid}, and FaceStudio~\cite{yan2023facestudio} utilize image $I$ as the target latent $Z$, with the facial region of $I$ serving as $C_i$. Conversely, Anydoor~\cite{chen2023anydoor}, and IP-Adapter directly employ the feature of image $I$ as $C_i$. In the learning phase of image reconstruction, this approach provides overly strong conditions for the diffusion model, which not only concentrates on facial features but also encompasses extraneous features such as the background and angles. This may result in the neglect of domain-invariant identity features.

\noindent\textbf{Random Reference Training Strategy.}
By directly applying the above method to video generation tasks, this strong conditioning can cause the video content to become heavily dependent on the semantic information of the reference image rather than focusing on its facial embedding. Character identity should exhibit domain invariance, implying that given images of the same individual from various angles and attire, video generation outcomes should be similar. Therefore, drawing inspiration from the Monte Carlo concept, we designed a random reference training methodology. This approach employs weakly correlated images with the current video sequence as the condition $C_j$, effectively decoupling the generated content from the reference images. Specifically, during training, we randomly select a reference image from the previously extracted face pool, as depicted in Figure~\ref{fig:mainfig}. By employing this Monte Carlo technique, the features from diverse reference images are averaged, reducing the influence of identity-invariant features. This transformation of the mapping from $(C,C_i)->z$ to $(C,C_j)->z$ diminishes the impact of extraneous features.

\noindent\textbf{Optimization with ID-Preserving Loss.}
In addition to the traditional diffusion loss function, we employ a reward function in our framework to promote identity-preserving learning. During the training process, we denoise the latent variables $z_t$, and after reaching a specific time step $\hat{t}$, the denoised latent $z_{0}$ is directly predicted from $z_{\hat{t}}$. Subsequently, we pass $z_{0}$ through the VAE decoder to obtain $X^{0}$. To measure the similarity between the reference face image and the generated content, we employ a face detection model to extract the face region from the generated content and use a face encoder to compute their similarity. This similarity is then used as a reward to update the model.

Given the face region of $X^{0}$ as $X^{f}$ and the condition image as $C_j$, the reward can be expressed as:
\begin{equation}
    \mathcal{R}_{id}(C_j,X^{f}) = \textit{cosSim}(\phi(C_j),\phi(X^{f}))
\end{equation}
where $\textit{cosSim(.)}$ represents the cosine similarity operator, and $\phi(.)$ denotes the face encoder.
By incorporating this identity-preserving reward, our approach can achieve superior identity-preserving capabilities. The overall optimization objective can be formulated as:
\begin{equation}
    \mathcal{L} = \mathcal{L}_{diff}+\lambda (1-\mathcal{R}_{id})
\end{equation}
where $1-\mathcal{R}_{id}$ is the identity-preserving loss and $\lambda$ is the hyper-parameter to balance the two losses.
\section{Experiment}
\subsection{Implementation details}
ID-Animator is compatible with various T2V generation backbones and here we employ the commonly-used AnimateDiff architecture~\cite{guo2023animatediff} for experiments. Our training dataset is processed by clipping to 16 frames, center cropping, and resizing to 512x512 pixels. During training, only the parameters of the face adapter are updated, while the pre-trained text-to-video model remains frozen. Our experiments are carried out on a single NVIDIA A100 GPU (80GB) with a batch size of 2. We load the pretrained weights of IP-Adapter and set the learning rate to 5e-5 for our trainable adapter. Furthermore, to enhance the generation performance using classifier-free guidance, we applied a 20\% probability of utilizing null-text embeddings to replace the original updated text embedding. Our training process is comprised of two stages. In the initial stage, we employ both the identity loss and diffusion loss to train the model, sampling only a single frame per video while setting the hyper-parameter $\lambda$ to 1. In the subsequent stage, we continue the training from the model obtained in the first stage, only utilized diffusion loss on 16-frame videos.
We trained our model for 1 epoch in the first stage and for 8 epochs in the second stage.

\subsection{Dataset and Evaluation Metrics}
We utilize a subset of the CelebV-text dataset as our training dataset, comprising 15k videos, and construct our identity-oriented dataset based on this foundation. Following the filtering of videos containing multiple faces, the final dataset employed for training contains 13k videos. To assess the performance of our methods, we opt for a comparative study with the IP-Adapter~\cite{ye2023ip} and AnimateDiff. To demonstrate the generalization ability of our methods, we utilize unsplash50~\cite{gal2024lcm} dataset as the testset and use prompts generated by GPT-4. 
We report the CLIP-I score, which measures the face structural similarity between the reference image and the generated video; the Dover score~\cite{wu2023exploring}, representing the overall quality of the generated video from both technical and aesthetic perspectives; the Motion Score~\cite{li2018vmaf}, assessing the diversity and variability of motion; the dynamic degree~\cite{huang2023vbench}, indicating the probability of generating motion-rich videos; and the Face Similarity~\cite{deng2019arcface}, which serves to measure facial feature similarity. We employ the CLIP encoder, which is pretrained on the LAION-Face dataset and is skilled at capturing facial structures. Given that CLIP and face encoder are trained on images, we calculate the CLIP-I, and face similarity for each frame and report the average score. The former three metrics are leveraged to evaluate the video quality, while the latter two serve to evaluate the identity preserving ability.
\begin{table}[!h]
\centering
\normalsize
 	\renewcommand{\tabcolsep}{3pt} 
  \caption{Quantitative comparison between the state-of-the-art methods. The best results are highlighted in bold, while the second-best results results are underlined.\label{tab1}}
\renewcommand{\arraystretch}{1.5}
\scalebox{0.8}{
\begin{tabular}{c|ccc|cc}
\hline
\multirow{2}{*}{Method}     & \multicolumn{3}{c|}{Video Quality}                                                    & \multicolumn{2}{c}{Identity Preservation}     \\ \cline{2-6} 
                            & \multicolumn{1}{c|}{Dover Score$\uparrow$} & \multicolumn{1}{c|}{Motion Score$\uparrow$} & Dynamic Degree$\uparrow$ & \multicolumn{1}{c|}{CLIP-I$\uparrow$} & Face Similarity$\uparrow$ \\ \hline
AnimateDiff~\cite{guo2023animatediff}                 & \multicolumn{1}{c|}{0.644}       & \multicolumn{1}{c|}{\underline{6.341}}        & \underline{0.380}          & \multicolumn{1}{c|}{-}      & -               \\ \hline
IP Adapter Plus Face~\cite{ipf}        & \multicolumn{1}{c|}{\underline{0.645}}       & \multicolumn{1}{c|}{3.799}        & 0.197          & \multicolumn{1}{c|}{\underline{0.805}}  & 0.266           \\ \hline
IP Adapter Face-ID Portrait~\cite{ipid} & \multicolumn{1}{c|}{0.588}       & \multicolumn{1}{c|}{5.480}        & 0.217          & \multicolumn{1}{c|}{0.587}  & \textbf{0.411}           \\ \hline
Ours                        & \multicolumn{1}{c|}{\textbf{0.714}}       & \multicolumn{1}{c|}{\textbf{8.008}}        & \textbf{0.539}          & \multicolumn{1}{c|}{\textbf{0.850}}  & \underline{0.316}           \\ \hline
\end{tabular}
}
\end{table}
\subsection{Quantitative Comparison}
The results of the quantitative experiment are presented in Table~\ref{tab1}. Our method surpassed the state-of-the-art methods across four metrics: CLIP-I, Dover Score, Motion Score, and Dynamic degree, achieving superior results. The IP Adapter Face ID Portrait employs arcface face embedding instead of clip embedding, which enhances the facial feature similarity but hinders the facial structural similarity. A direct application of the IP Adapter to video models results in diminished video quality and relatively static motion, as evidenced by the dover score, motion score and the dynamic degree. Additionally, our methods also surpassed the AnimateDiff on human related prompts, demonstrating our superior human related video generation ability.
\begin{figure}[h]
    \centering
    \includegraphics[width=\linewidth]{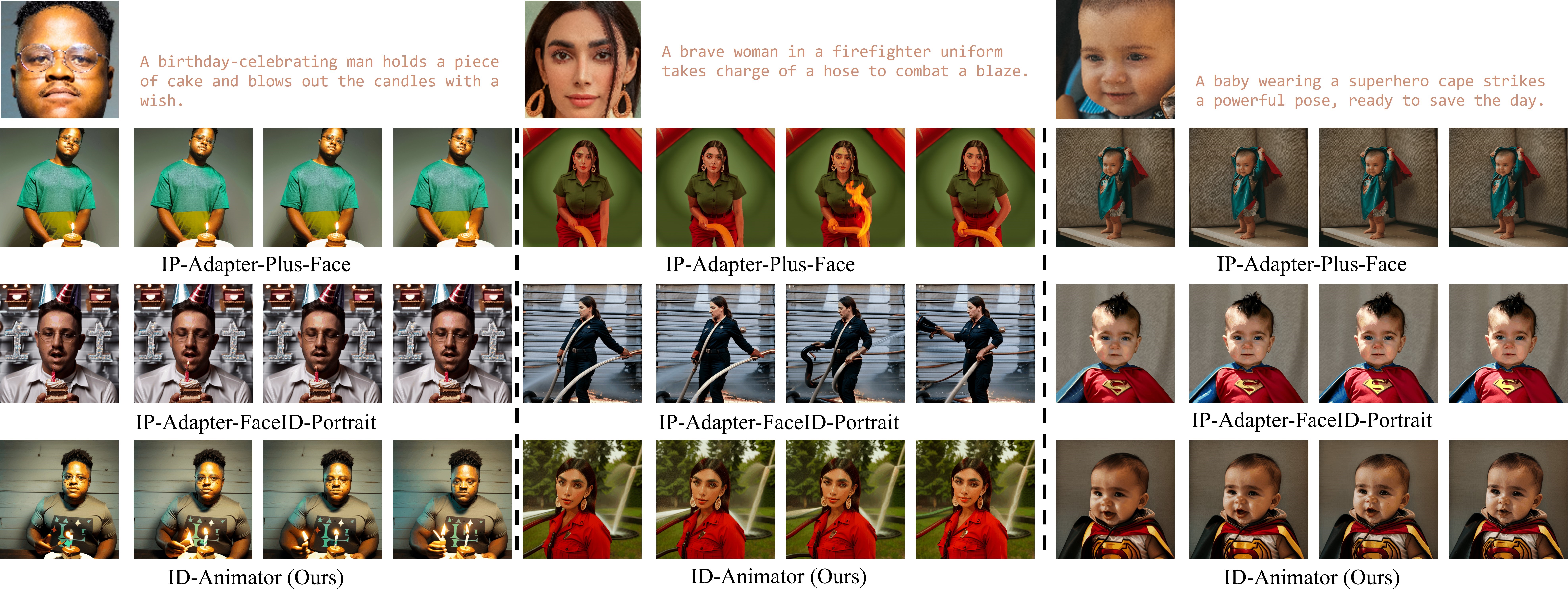}
    \caption{Comparison between our methods and previous methods on three ordinary individuals.}
    \label{fig:cmp2}
\end{figure}
\subsection{Qualitative Comparison}
We choose three images of ordinary individuals as test cases, with the images being sampled from unseen testing set. We randomly generated three prompts from LLM, maintaining consistency with human language style.
As depicted in Figure~\ref{fig:cmp2}, it is shown that our results can better preserve the identity information of the given reference image over other methods, no matter for man, woman or kids, showing the identity fidelity and generalization capacity of our method. In contrast, the face generated by IP-Adapter-Plus-Face shows a certain level of deformation, whereas the IP-Adapter-FaceID-Portrait model is deficient in facial structural information, resulting in a diminished similarity between the generated outputs and the reference image. 

\subsection{Applications}
In this section, we showcase the potential applications of our model, encompassing recontextualization, alteration of age or gender, ID mixing, and integration with ControlNet or community models~\cite{civitai} to generate highly customized videos.
\subsubsection{Recontextualization}
\begin{figure}
    \centering
    \includegraphics[width=\linewidth]{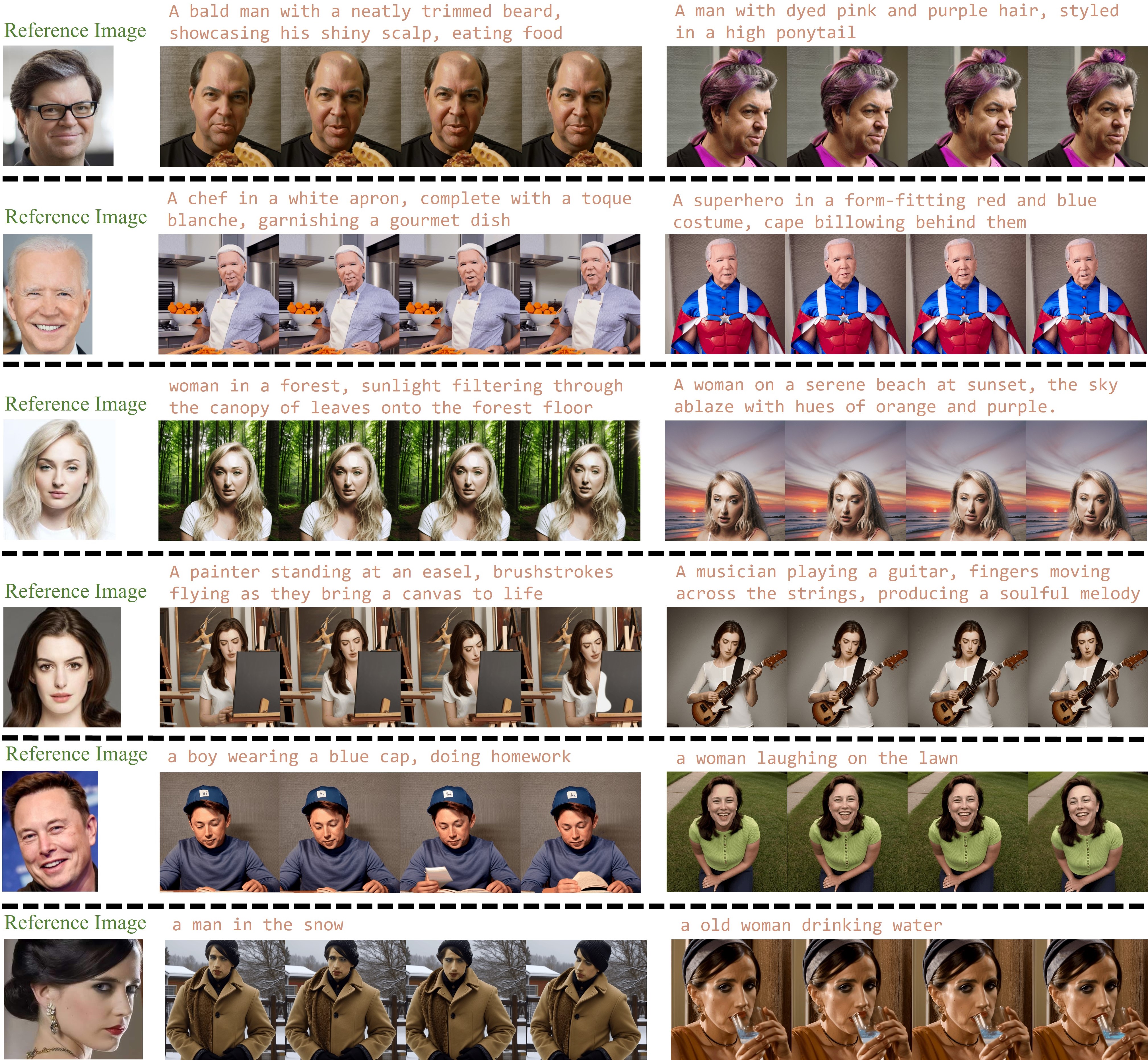}
    \caption{From top to bottom, our model showcases its ability to recontextualize various elements in an reference image, including human hair, clothing, background, actions, age, and gender.}
    \label{fig:recon}
\end{figure}

Given a reference image, our model is capable of generating ID fidelity videos and changing contextual information. The contextual information of characters can be tailored through text, encompassing attributes such as features, hair, clothing, creating novel character backgrounds, and enabling them to execute specific actions. As illustrated in Figure~\ref{fig:recon}, we supply reference images and text, and the outcomes exhibit the robust editing and instruction-following capacities of our model.

As depicted in the figure~\ref{fig:recon}, from top to bottom, we exhibit the model's proficiency in altering character hair, clothes, background, executing particular actions, and changing age or gender.
\begin{figure}
    \centering
    \includegraphics[width=\linewidth]{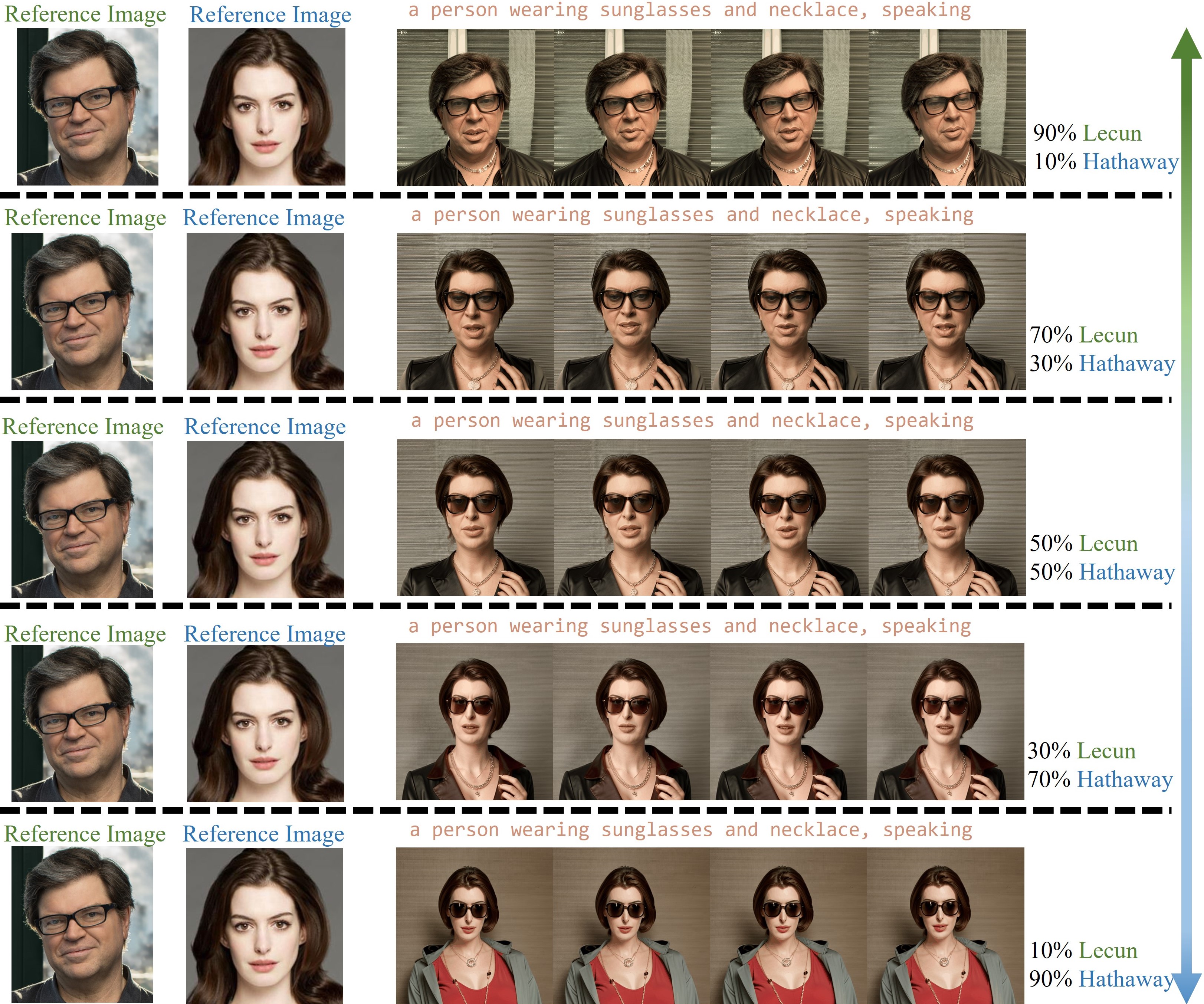}
    \caption{The figure illustrates our model's capability to blend distinct identities and create identity-specific videos.}
    \label{fig:id-mixing}
\end{figure}
\subsubsection{Identity Mixing}
The potential of our model to amalgamate different IDs is showcased in the figure~\ref{fig:id-mixing}. Through the blending of embeddings from two distinct IDs in varying proportions, we have effectively combined features from both IDs in the generated video. This experiment substantiates the proficiency of our face adapter in learning facial representations.

\begin{figure}
    \centering
    \includegraphics[width=\linewidth]{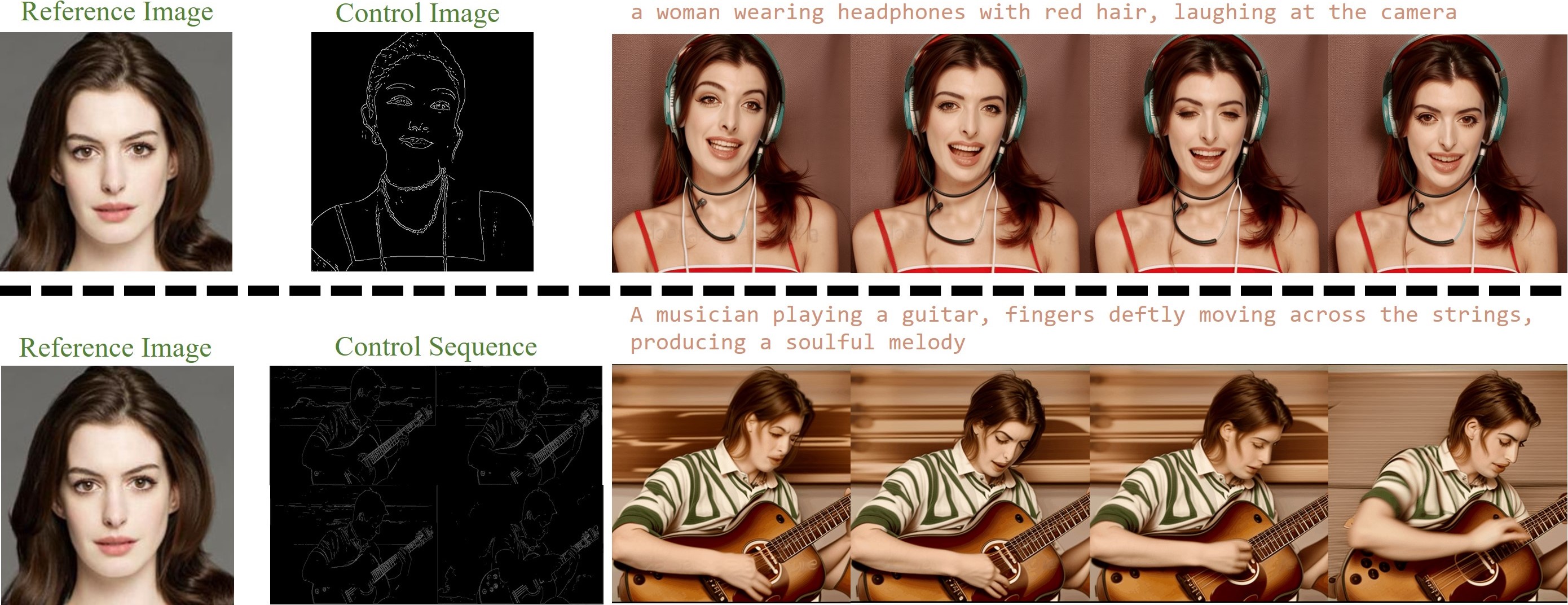}
    \caption{Our model can combine with ControlNet to generate ID-specific videos.}
    \label{fig:control}
\end{figure}
\subsubsection{Combination with ControlNet}
Furthermore, our model demonstrates excellent compatibility with existing fine-grained condition modules, such as ControlNet~\cite{zhang2023adding}. We opted for SparseControlNet~\cite{guo2023sparsectrl}, trained for AnimateDiff, as an additional condition to integrate with our model. As illustrated in Figure~\ref{fig:control}, we can supply either single frame control images or multi-frame control images. When a single frame control image is provided, the generated result adeptly fuses the control image with the face reference image. In cases where multiple control images are presented, the generated video sequence closely adheres to the sequence provided by the multiple images. This experiment highlights the robust generalization capabilities of our method, which can be seamlessly integrated with existing models.
\begin{figure}
    \centering
    \includegraphics[width=\linewidth]{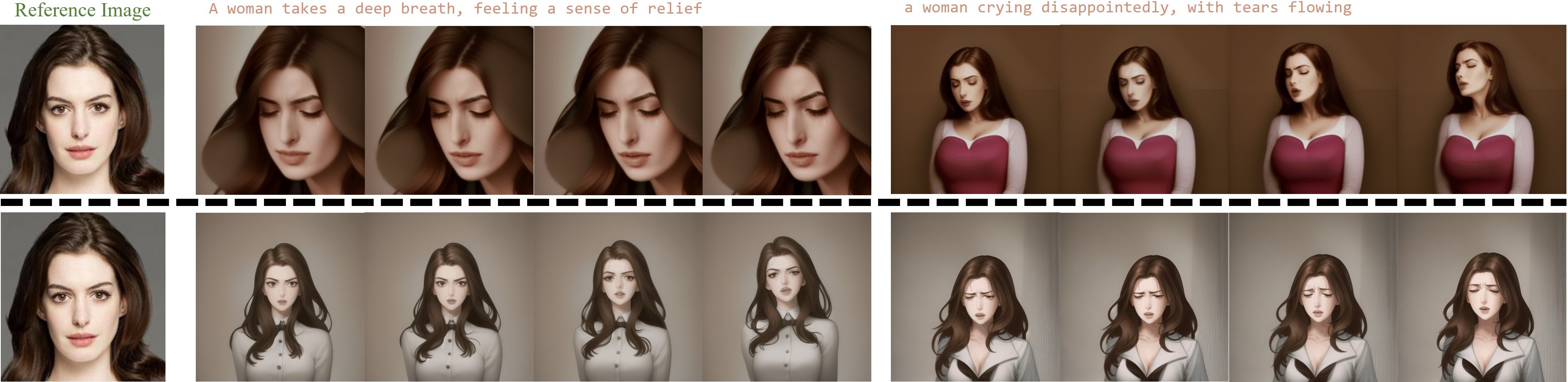}
    \caption{From the top to bottom, we visualize the inference results with Lyriel and Raemuxi model weights.}
    \label{fig:commu}
\end{figure}
\subsubsection{Inference with Community Models}
We assessed the performance of our model using the \href{https://civitai.com/} {Civitai} community model, and our model continues to function effectively with these weights, despite never having been trained on them. The selected models include \href{https://civitai.com/models/22922/lyriel}{Lyriel} and \href{https://civitai.com/models/113362/raemumix}{Raemumxi}. As depicted in Figure~\ref{fig:commu}, the first row presents the results obtained with the Lyriel model, while the second row showcases the outcomes achieved using the Raemuxi model. Our method consistently exhibits reliable facial preservation and motion generation capabilities.

\subsection{Ablation Experiment}
To investigate the efficacy of our proposed method, we carry out ablation experiments for ID-Animator. We first study the effectiveness of the unified captioning technique, the random reference training strategy, as well as the  ID-preserving objective. In addition, we also demonstrate the recontextualization and identity mixing ability of our model. Following the literature~\cite{hu2024ella}, we random sample a subset from the unslash50 dataset for ablation study due to computation resources issue.

\textbf{On the Effectiveness of ID Preserving Loss:}
In the first part of ablation experiments, we remove the ID-preserving objective. As demonstrated by the Table~\ref{tab2}, the removal of the ID-preserving loss leads to a decrease in the model's facial feature similarity and the CLIP-I metrics, validating the effectiveness of ID-preserving objective. Furthermore, the motion score and dynamic degree suffer from a notable degradation. 

\textbf{On the Effectiveness of Random Reference Training:} In the second set of ablation experiments, we remove the random reference training strategy and utilize a fixed reference image as the facial region in the first frame of the video. As shown in the fourth row of Table~\ref{tab2}, upon removal of the random reference training strategy, we observe a decline in video quality and the metrics of CLIP-I. Without random reference training strategy, the model is inevitably confused by the irrelevant semantic information in the reference image while neglecting the identity-relevant embeddings, leading to inferior performance for generated human videos.

\textbf{On the Effectiveness of Unified Captioning on Human Videos:} In the third ablation study, we utilized the original captions from the CelebV dataset for model training. Due to the limited diversity in the available captions, we noticed a significant decrease in both motion and dover scores of the generated videos. Without the newly-provided unified captions, the model struggles to produce motion-enriched videos given human-like style text inputs, resulting in poor video quality. 
\subsection{Prompt List for Evaluation}
In order to assess our model's video generation capabilities, we follow the previous method~\cite{xiao2023fastcomposer}, employing four categories of prompts: Accessory, Style, Context, and Action. Distinct from prior methods, we leveraged GPT4 to rewrite these prompts, enhancing their complexity and aligning them with human language styles. We utilized a total of 50 prompts, with the Unsplash50 dataset comprising 50 unique IDs. For each ID, we generated 50 videos using the 50 prompts, resulting in a total of 2,500 videos for each method. The prompt content, displayed in the Table~\ref{tab:prompt}, showcases the diversity and human-like style.
\begin{table}[!h]
\centering
\huge
 	\renewcommand{\tabcolsep}{3pt} 
  \caption{The prompt is used in the evaluation procedure of the adapter model. This prompt is rewritten by GPT-4 and is highly aligned with human style.\label{tab:prompt}}
\renewcommand{\arraystretch}{2.5}
\resizebox{\linewidth}{!}{
\begin{tabular}{c|ccccc|c}
\cline{1-2} \cline{6-7}
Category                    & Prompt                                                                                                           &  &  &  & Category                  & Prompt                                                                                                       \\ \cline{1-2} \cline{6-7} 
\multirow{10}{*}{Accessory} & A \{class\_token\} donning a red hat greets others with a friendly wave.                                         &  &  &  & \multirow{10}{*}{Cotenxt} & A curious \{class\_token\} in the heart of the jungle examines a map to navigate the dense foliage.          \\
                            & A \{class\_token\} in a Santa hat enjoys a warm cup of hot cocoa during the holiday season.                      &  &  &  &                           & An adventurous \{class\_token\} in the snow constructs a snowman with joy and laughter.                      \\
                            & A \{class\_token\} wrapped in a vibrant rainbow scarf snaps a cheerful selfie.                                   &  &  &  &                           & A sun-loving \{class\_token\} on the beach diligently applies sunscreen to protect their skin.               \\
                            & A sophisticated \{class\_token\} wearing a black top hat and monocle peruses the daily newspaper.                &  &  &  &                           & A \{class\_token\} strolls along a cobblestone street, sketching the charming surroundings.                  \\
                            & A culinary \{class\_token\} dressed in a chef's attire expertly seasons a delicious dish.                        &  &  &  &                           & A focused \{class\_token\} sits on pink fabric, skillfully sewing a button onto a garment.                   \\
                            & A brave \{class\_token\} in a firefighter uniform takes charge of a hose to combat a blaze.                      &  &  &  &                           & A patient \{class\_token\} on a wooden floor assembles a puzzle, piece by piece.                             \\
                            & A vigilant \{class\_token\} in a police outfit communicates with colleagues over a radio.                        &  &  &  &                           & A busy \{class\_token\} with a cityscape in the background hails a taxi to their next destination.           \\
                            & A stylish \{class\_token\} sporting pink glasses browses the latest trends on a tablet.                          &  &  &  &                           & A nature-loving \{class\_token\} with a majestic mountain in the background takes a deep, refreshing breath. \\
                            & A creative \{class\_token\} wearing a bright yellow shirt skillfully paints a masterpiece.                       &  &  &  &                           & A \{class\_token\} with a quaint blue house in the background tends to their vibrant garden.                 \\
                            & A mystical \{class\_token\} adorned in a purple wizard outfit casts a spell with a flourish.                     &  &  &  &                           & A reflective \{class\_token\} on a purple rug in a serene forest writes their thoughts in a journal.         \\ \cline{1-2} \cline{6-7} 
\multirow{10}{*}{Style}     & A thought-provoking painting of a \{class\_token\} in Banksy's street art style, cleverly spray painting a wall. &  &  &  & \multirow{10}{*}{Action}  & A confident \{class\_token\} riding a horse adjusts their hat while maintaining control.                     \\
                            & A \{class\_token\} captured in the vivid style of Vincent Van Gogh, gently picking flowers from a field.         &  &  &  &                           & A sociable \{class\_token\} holding a glass of wine raises a toast to celebrate with friends.                \\
                            & A lively graffiti painting of a \{class\_token\} passionately strumming a guitar.                                &  &  &  &                           & A birthday-celebrating \{class\_token\} holds a piece of cake and blows out the candles with a wish.         \\
                            & A serene watercolor painting of a \{class\_token\} gracefully holding an umbrella in the rain.                   &  &  &  &                           & An intellectual \{class\_token\} giving a lecture adjusts their glasses for a clearer view.                  \\
                            & A Greek marble sculpture of a \{class\_token\} reflecting on their own beauty.                                   &  &  &  &                           & A studious \{class\_token\} reading a book turns a page, eager to continue the story.                        \\
                            & A captivating street art mural of a \{class\_token\} taking a photo of a bustling city scene.                    &  &  &  &                           & A green-thumbed \{class\_token\} tends to their backyard garden, pruning plants with care.                   \\
                            & A nostalgic black and white photograph of a \{class\_token\} lighting a cigarette in a quiet moment.             &  &  &  &                           & A \{class\_token\} cooking a meal stirs a pot, ensuring the flavors meld together perfectly.                 \\
                            & A pointillism painting of a \{class\_token\} playfully interacting with a delicate butterfly.                    &  &  &  &                           & A determined \{class\_token\} at the gym wipes their brow after an intense workout.                          \\
                            & A traditional Japanese woodblock print of a \{class\_token\} pouring tea with elegance.                          &  &  &  &                           & A responsible \{class\_token\} walks their dog, holding the leash to ensure their pet's safety.              \\
                            & A bold street art stencil of a \{class\_token\} writing a powerful message for all to see.                       &  &  &  &                           & A \{class\_token\} baking cookies takes a taste of the dough, ensuring it's just right.                      \\ \cline{1-2} \cline{6-7} 
\end{tabular}}
\end{table}

\begin{table}[!h]
\centering
 	\renewcommand{\tabcolsep}{3pt} 
  \caption{Ablation experiment results on unsplash50 dataset. The best results are highlighted in bold, while the second-best results results are underlined.\label{tab2}}
\renewcommand{\arraystretch}{1.5}

\resizebox{\linewidth}{!}{
\begin{tabular}{c|c|c|c|ccc|cc}
\hline
\multirow{2}{*}{Config} & \multirow{2}{*}{ID Loss} & \multirow{2}{*}{Random Reference Training} & \multirow{2}{*}{Unified Captioning} & \multicolumn{3}{c|}{Video Quality}                                                    & \multicolumn{2}{c}{Identity Preservation}     \\ \cline{5-9} 
                        &                          &                                   &                                  & \multicolumn{1}{c|}{Dover Score$\uparrow$} & \multicolumn{1}{c|}{Motion Score$\uparrow$} & Dynamic Degree$\uparrow$ & \multicolumn{1}{c|}{CLIP-I$\uparrow$} & Face Similarity$\uparrow$ \\ \hline
(I)                     & x                        & \checkmark                                 & \checkmark                                & \multicolumn{1}{c|}{\underline{0.730}}       & \multicolumn{1}{c|}{\underline{5.694}}        & \underline{0.324}          & \multicolumn{1}{c|}{0.750}  & 0.292           \\ \hline
(II)                    & \checkmark                        & x                                 & \checkmark                                & \multicolumn{1}{c|}{0.700}       & \multicolumn{1}{c|}{5.318}        & 0.182          & \multicolumn{1}{c|}{\underline{0.760}}  & \textbf{0.321}           \\ \hline
(III)                   & \checkmark                        & \checkmark                                 & x                                & \multicolumn{1}{c|}{0.679}       & \multicolumn{1}{c|}{4.158}        & 0.158          & \multicolumn{1}{c|}{0.740}  & 0.300           \\ \hline
Ours                    & \checkmark                        & \checkmark                                 & \checkmark                                & \multicolumn{1}{c|}{\textbf{0.739}}       & \multicolumn{1}{c|}{\textbf{7.797}}        & \textbf{0.507}          & \multicolumn{1}{c|}{\textbf{0.768}}  & \underline{0.315}           \\ \hline
\end{tabular}
}
\end{table}

\begin{figure}
    \centering
    \includegraphics[width=0.9\linewidth]{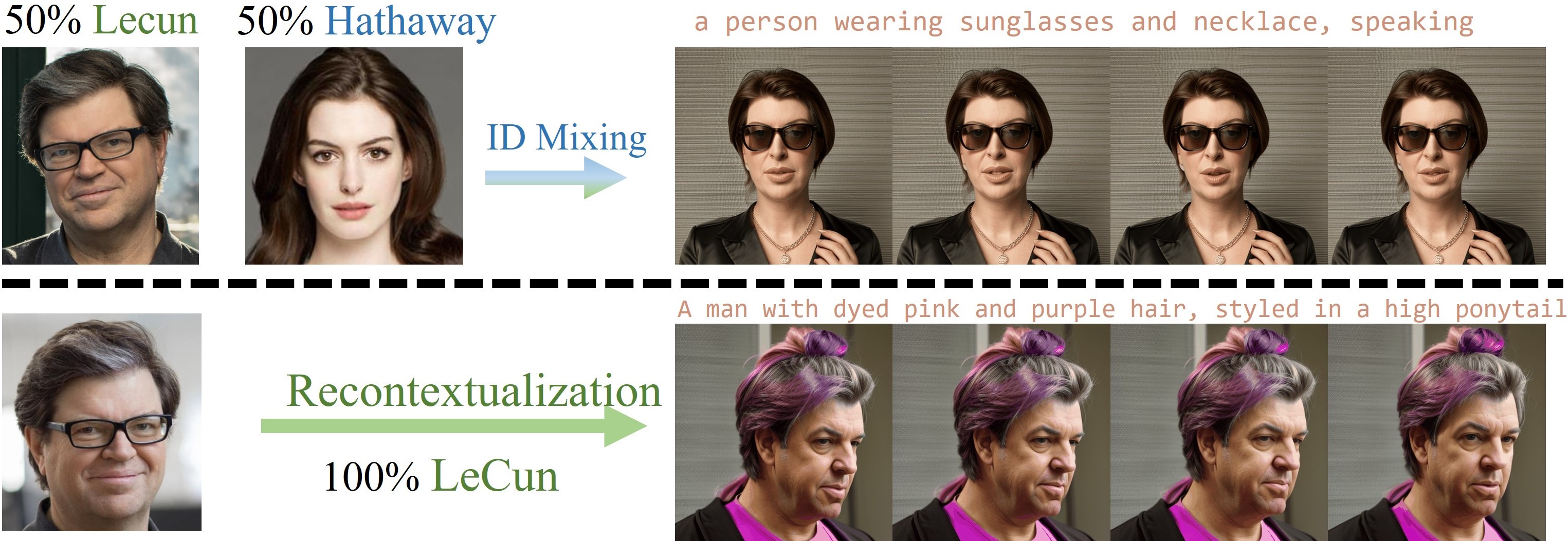}
    \caption{Demonstration for the recontextualization and identity mixing ability of our methods.}
    \label{fig:mixing}
\end{figure}

\textbf{Recontextualization and Identity Mixing Capacity:} Apart from the traditional video generation ability, we further validate the recontextualization and the identity mixing ability of our method. As illustrated in the Figure~\ref{fig:mixing}, we observe that our method is able to manipulate the facial attributes (e.g., hair style and color) of given identity via text condition, showing the recontextualization capacity. Moreover, when given multiple facial images from different person as input, our model is further able to generate videos that seamlessly mixes these individual facial identities. These results show the generalization capacity and extendability of our method in real-world applications.
\begin{figure}
    \centering
    \includegraphics[width=\linewidth]{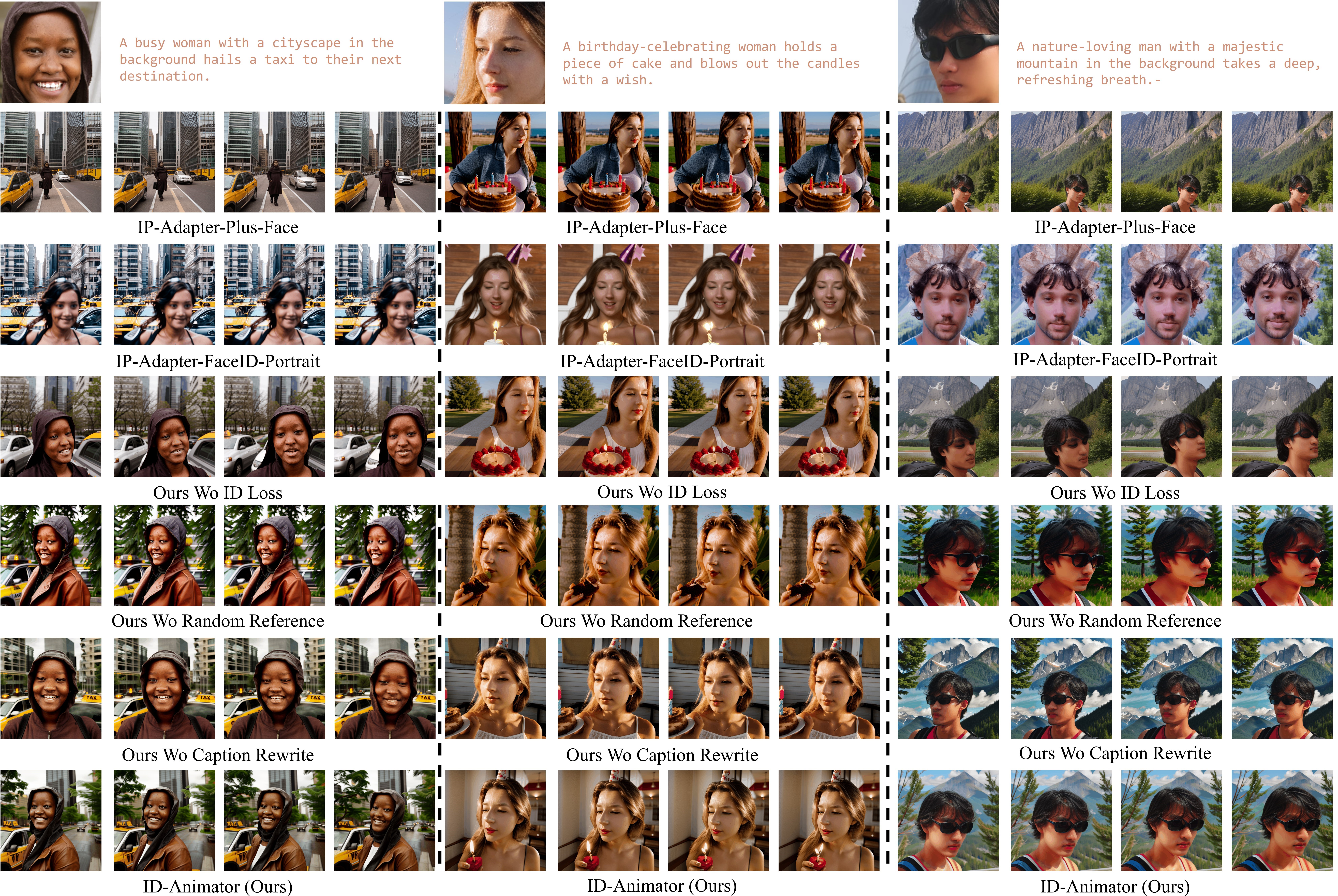}
    \caption{The figure demonstrates the effectiveness of our proposed methods.}
    \label{fig:ablvisual}
\end{figure}

\textbf{Visual Results of Ablation Experiment}
Ee showcase additional experimental results, encompassing visualizations of ablation studies and comparisons with IP adapters, as depicted in the Figure~\ref{fig:ablvisual}. In the left column of the figure, it is evident that the IP adapter Plus Face produces an excessively small face, leading to a blurry and unrecognizable generated output. Meanwhile, the IP adapter FaceID portrait lacks the necessary facial structural information, resulting in a substantial discrepancy between the generated output and a real person's appearance. When contrasted with the ablation study, we observe that the absence of ID loss contributes to a decline in facial similarity, while the lack of random reference training causes the generated outputs to be nearly static. From the middle column of the image, it becomes apparent that faces generated by other methods appear obese, consequently compromising the original facial structure information. The third column of the image presents analogous results.
Theses visual results demonstrate the effectiveness of our methods which combining the random reference training methods, the dataset reconstruction pipeline and the ID-preserving loss function.

\section{Conclusion}
In this research, our primary goal is to achieve ID-specific content generation in text-to-video (T2V) models. To this end, we introduce a ID-Animator framework to drive T2V models in generating ID-specific human videos using ID images. We facilitate the training of our ID-Animator by constructing an ID-oriented dataset based on publicly available resources, incorporating unified caption generation and face pool construction. 
Moreover, we develop a random reference training method to minimize ID-irrelevant content in reference images and utilized an ID preservation loss function to encourage ID preservation learning, thereby directing the adapter's focus towards ID-related features. Our extensive experiments demonstrate that our ID-Animator generates stable videos with superior ID fidelity compared to previous models. 

\bibliographystyle{plain}
\bibliography{nips}

\end{document}